\documentclass[review,12pt]{elsarticle}
\usepackage{amsmath,amssymb,latexsym}
\usepackage{lineno,hyperref}
\usepackage{multirow}
\usepackage{graphicx}
\usepackage{adjustbox}
\usepackage{subfigure}
\usepackage{array}
\usepackage{booktabs}
\usepackage{setspace}
\usepackage{caption}
\graphicspath{{./figs/}}
\journal{Journal of \LaTeX\ Templates}

\begin{document}
%\begin{spacing}{2.0}
\begin{frontmatter}

\title{Low-rank features based double transformation matrices learning for image classification}

\author{Yu-Hong Cai}
\author{Xiao-Jun Wu$^*$}
\author{Zhe Chen}

\address{School of Artificial Intelligence and Computer Science, Jiangnan University, Wuxi 214122, China}

\begin{abstract}
	Linear regression is a supervised method that has been widely used in classification tasks. In order to apply linear regression to classification tasks, a technique for relaxing regression targets was proposed. However, methods based on this technique ignore the pressure on a single transformation matrix due to the complex information contained in the data. A single transformation matrix in this case is too strict to provide a flexible projection, thus it is necessary to adopt relaxation on transformation matrix. This paper proposes a double transformation matrices learning method based on latent low-rank feature extraction. The core idea is to use double transformation matrices for relaxation, and jointly projecting the learned principal and salient features from two directions into the label space, which can share the pressure of a single transformation matrix. Firstly, the low-rank features are learned by the latent low rank representation (LatLRR) method which processes the original data from two directions. In this process, sparse noise is also separated, which alleviates its interference on projection learning to some extent. Then, two transformation matrices are introduced to process the two features separately, and the information useful for the classification is extracted.  Finally, the two transformation matrices can be easily obtained by alternate optimization methods. Through such processing, even when a large amount of redundant information is contained in samples, our method can also obtain projection results that are easy to classify. Experiments on multiple data sets demonstrate the effectiveness of our approach for classification, especially for complex scenarios.
\end{abstract}

\begin{keyword}
	Least square regression\sep Low-rank representation\sep Feature extraction\sep Double transformation matrices learning
\end{keyword}

\end{frontmatter}

%\linenumbers

\section{Introduction}
Feature extraction \cite{zheng2006nearest,zheng2006reformative,wu2004new} is a key ingredient in pattern classification, as well as in image fusion \cite{luo2016novel,luo2017image,li2017multi,li2020nestfuse} and many other machine vision tasks\cite{li2011no,chen2018new,sun2019effective,wang2003initial,sun2011quantum}. Recently, least square regression (LSR) based methods have been widely used for tasks in the fields of pattern recognition and computer vision (\cite{LRforspeech};\cite{roburegree}). LSR is a simple but efficient tool for data analysis because of its mathematical tractability. Many variants of LSR have been developed for different scenes, such as weighted LSR (\cite{weightlsr}), partial LSR (\cite{partiallsr}), kernel LSR (\cite{kernellsr}) and other extensions.

The conventional LSR aims to find a transformation matrix that can transform samples into the corresponding label matrix perfectly. Given a sample $x_i\in R^{m}$, the objective function is given as follows:
\begin{equation}
\min\limits_W\left\|y_i-Wx_i\right\|_2^2+\lambda\left\|W\right\|_F^2
\end{equation}

\noindent where $W\in R^{c\times m}$ is the transformation matrix, and $y_i\in R^{c}$ is the label matrix of data $x_i$, which is defined that if $x_i$ belongs to the $k$th class $(k=1,...,c)$, the $k$th element of $y_i$ is one and all other elements are zero. Sparse constraints can be added to the transformation, such as the $l_1$-norm or the $l_{2,1}$-norm (\cite{sparsity}), allowing the transformation matrix to choose more discriminative features. However, for multi-class classification in practice, strict $0-1$ label matrix is too rigid to get a discriminative transformation matrix $W$ for classification tasks. A slack structure can be more flexible and provides more discriminative results (\cite{SB2DL}). Xiang et al. (\cite{dlsr}) proposed a technique which is called $\epsilon$-dragging to force the regression target moving along opposite direction, and formulated a discriminative LSR (DLSR) model.

In addition to the most representative DLSR model, based on the relaxed label matrix, many works have been developed. Wang and Zhang et al. imposed an $l_1$-norm constraint (\cite{msdlsr}) on the $\epsilon$-dragging matrix to control the margin of DLSR. Chen and Wu et al. imposes low-rank constraint on the relaxed label matrices class-wisely (\cite{lrdlsr}). Zhang et al. proposed retargeted LSR (ReLSR) (\cite{relsr}) which introduces a margin constraint that forces the margin between true and false regression targets to be larger than 1. So ReLSR does not use the $\epsilon$-dragging but applies the constraint to get the suitable regression targets from the data. Based on the works in DLSR and ReLSR, Wang and Pan (\cite{grelsr}) pointed out that DLSR is a special case of ReLSR with the translation value being 0, and proposed groupwise retargeted LSR (GReLSR). To restrict the translation values of ReLSR, an additional groupwise regularization is utilized to force samples from the same class to have the same translation value in GReLSR. To further exploit correlations among samples, the graph structure is introduced into the LSR model (\cite{drlsr}). Fang et al developed a regularized label relaxed (RLR) (\cite{rlr}) linear regression method with the class compactness graph which ensures that the samples of each class can be kept close after they are transformed.

All the above methods draw their attention to the regression target and the transformation matrix, and there are few works about the data to be transformed. The data we use in the classification task usually contains a lot of mixed information, such as class-specific information, inter-class relationship information, intra-class variations and so on (\cite{structurenoise};\cite{srnmr}). In such cases, the reason why the good classification result is hard to be obtained may be due to the complexity of the information of the data. In robust latent subspace learning (RLSL) (\cite{rlsl}), a subspace learning technique is imposed, so that a more compact data representation obtained from the robust subspace is used to be transformed. During the learning of subspace, sparse noise part is separated and only the clean data representation is selected to learn the transformation matrix.

RLSL model attempts to learn compact representation of data through subspace projection to obtain more discriminative feature information. Although the effect of sparse noise is excluded in this process, they do not pay attention to the various types of information contained in the data, and the compact representation still contains mixed information. For such classification tasks, the transformation matrix projects complex data information into the label space, which may be too strict for a single transformation matrix. However, in these methods mentioned above, they only use a single transformation matrix to transform samples. Based on similar motivations, Han and Wu et al. proposed a double relaxed regression (DRR) method (\cite{drr}). They pointed out that such a single transformation may be too strict to provide more freedom for learning better margins. To this end, DRR introduces another transformation matrix to share part of responsibility of original single transformation matrix.

One problem in DRR is that another transformation matrix introduced is not used to project the data to the label matrix, but only exists in the constraint and affects the projection matrix indirectly. In this way, the intra-class differences in the original data due to expressions, illuminations and poses are still handled by a single transformation matrix during the projection process. These differences may not be well eliminated to minimize the distance between the transformed data from the same class during the projection.

The methods mentioned above are very useful and have a good classification effect. However, they all focused on the relaxation of the regression target, without paying attention to the complex information contained in the raw data. It is thus neglected that the transformation matrix responsible for extracting the classification features of the data does not perform the task well because of these complex mixed information in the sample data. In order to process the various information in the original samples that is advantageous for classification, we may need to introduce two or more transformation matrices to process this information separately. Based on these views mentioned above, in this paper, we propose a new learning method for transformation matrices, which deals with feature extraction and regression tasks separately. First, the raw data is processed by LatLRR, and the principal and salient features are extracted from the column space and the row space respectively. Then two transformation matrices are introduced to process the two features separately, then the column and row space features are jointly projected onto the label matrix by the corresponding transformation matrix. By adopting two transformation matrices, the regression task is separated into two parts, which further relaxes the transformation matrix learning and eases the responsibility of transformation matrices to some extent. In brief, the proposed method has the following properties.

\begin{enumerate}
	\item We propose a novel transformation matrices learning method based on low-rank feature extraction and analysis of transformation matrices in least square regression.
	
	\item The low-rank feature extraction is used to decompose the original data into principal, salient information and sparse noise respectively. So that, the complex information in source data is decomposed into three parts. In the subsequent transformation matrices learning, sparse noise is not used, and only clean principal (column space) and salient (row space) feature parts are used for projection.
	
	\item For the principal and salient features, we apply corresponding two transformation matrices to process these two features separately and project the features extracted from column and row space of the source data to corresponding regression target jointly. Therefore, the task of projecting complex data is shared by two transformation matrices.
	
	\item In the classification, we use the low rank projection method to learn the projection matrix from the original data to the corresponding feature space, so that the column space features of the test samples can be quickly obtained for classification.
\end{enumerate}

The remainder of this paper is arranged as follows. We introduce some notations and the related works in Section 2. Then, we expound our motivation and the proposed methods in Section 3 followed with its optimal solution. The experiments and analysis are presented in Section 4. Section 5 concludes this paper.

\section{Related Work}\label{sec2}
This section briefly introduces the least square regression methods and low-rank representation based feature extraction methods.

\subsection{Least Square Regression Methods}
Given a data set $X$ and corresponding label matrix $Y$, the conventional LSR aims to learn a transformation matrix that can transform the data into their corresponding labels. the objective function of LSR is given as follow:
\begin{equation}
\min\limits_W\left\|Y-WX\right\|_F^2+\lambda\left\|W\right\|_F^2
\end{equation}

\noindent where $W\in \mathcal{R}^{c\times m}$ is the transformation matrix, and $Y=[y_1,...,y_n ]\in \mathcal{R}^{c\times n}$ is the label matrix of data $X\in \mathcal{R}^{m\times n}$.

LSR is simple but efficient, and its closed form solution can be obtained directly by solving linear equations, but LSR is not suitable for direct use in multi-class classification tasks.

\subsection{LRR-Based Feature Extraction}
Low-rank representation (LRR) (\cite{lrr}) model assumes that data are approximately represented by samples from multiple low-dimensional subspaces. In general, LRR aims at finding the lowest-rank representation that can represent the data as linear combinations of the basis in a given dictionary $D$:
\begin{equation}
\min\limits_{Z,E} rank(Z)+\lambda\left\|E\right\|_1\quad s.t. X=DZ+E
\end{equation}

\begin{figure}[htp]
	\centering
	%	\small
	\includegraphics[scale=0.5]{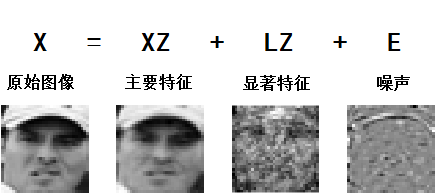}
	\\
	\includegraphics[scale=0.5]{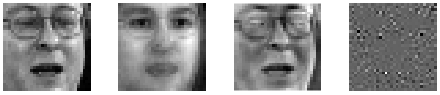}
	\caption{Some samples of using LatLRR to decompose the raw data. }
	\label{Fig.1}
\end{figure}
where every column of $D$ is a base which is learned from training sets, and usually the training data itself is usually used as the dictionary i.e. set $D=X$. Matrix $E$ denotes the sparse noise or error item. It is hard to solve the formulation above, and a convex relaxation is introduced to solve the problem:
\begin{equation}
\min\limits_{Z,E} \left\|Z\right\|_*+\lambda\left\|E\right\|_1\quad s.t. X=DZ+E
\end{equation}

However, when the training sample are insufficient, $Z$ probably gets the feasible solution and thus the performance of LRR model is reduced. To deal with the problem, except the observed data i.e. training data $X$, LatLRR \cite{latlrr} puts the hidden data or the unobserved data into consideration:
\begin{equation}
\min\limits_{Z,E} \left\|Z\right\|_*+\lambda\left\|E\right\|_1\quad s.t. X=[X_O,X_H]Z+E
\end{equation}
where $X_O$ denotes the observed data and $X_H$ denotes the hidden data which are unobservable. Based on the theorems mentioned in (\cite{lrr}), the objective function of LatLRR is formulated as:
\begin{equation}\label{latlrr}
\min\limits_{Z,L,E} \left\|Z\right\|_*+\left\|L\right\|_*+\lambda\left\|E\right\|_1\quad s.t. X=XZ+LX+E
\end{equation}

where $Z$ and $L$ are all low-rank matrices, and respectively reconstruct a data matrix $X$ from two directions: column and row. And $XZ$ can be regarded as the principal features while $LX$ as the salient features. On the one hand, the principal features $XZ$ extracted from column space of the raw data, which is expanded by columns of the raw data i.e. the given training samples, so that the principal features $XZ$ contain the relationship between samples. On the other hand, the salient features are derived from the row view, which contains the most critical information in each sample.

\section{Proposed Method}
In this section, we mainly introduce the motivation and the formulation of the proposed method and the corresponding optimization process.

\subsection{Formulation}
As mentioned before, when performing classification tasks, the sample data contains various information such as class specific information, association information between samples, intra-class variations, and inter-class differences. In this case, a single transformation matrix does not provide a flexible projection to eliminate the negative effect. At the same time, the most direct way to extract the information from the original data is to use the feature extraction method to preprocess the original data. Inspired by these, in order to process complex information in the raw data $X$, we use LatLRR to perform a pre-processing on the raw data. As described in the section (\ref{sec2}), LatlRR extracts two corresponding feature salient features and principal features from the original data $X$ from two directions: column and row. The features extracted in these two directions contain the key information of the sample itself and the association between the samples. The learning model is given by: 
\begin{equation}
\min\limits_{Z,L,E} \left\|Z\right\|_*+\lambda_1\left\|L\right\|_*+\lambda_2\left\|E\right\|_1   
s.t. X=XZ+LX+E
\end{equation}
%\begin{equation}
%s.t. X=XZ+LX+E
%\end{equation}

where $\lambda_1$ is a parameter that balances the principal feature and the salient feature, and $\lambda_2$ is a penalty parameter for sparse noise. Fig.\ref{Fig.1} shows the features extracted by LatLRR. The sparse noise term contains a large amount of random noise. On the one hand, it does not contain class-specific information, on the other hand it increases the burden of the transformation matrix, which does not help the classification regression task. So we only use the clean two feature parts for the next work. In general, in order to use the information in the feature to classify, the usual approach is to learn a transformation matrix to project feature information to the classification output, as is the case with most regression-based methods. Thus, by only seleting the clean part of the raw data to learn the transformation matrix $W$, we propose the following formulation:
\begin{equation}
\min\limits_{W}\left\|Y-W(XZ+LX)\right\|_F^2+\lambda\left\|W\right\|_F^2
\end{equation}

Although the sparse noise term is removed, there will still be differences in actual samples of the same type. And as mentioned in (\cite{drr}), a single transformation matrix is too strict for complex classification tasks to learn a better margin. Moreover, a single transformation matrix does not make good use of the information contained in the extracted features. In other words, in such task, the problem to simultaneously process features extracted from two different angles, is too complicated for a single projection matrix.

In order to make full use of the information contained in these two features for classification, we introduce two classifiers $W_1$ and $W_2$ to process the two features separately and project the corresponding feature information into the label space. Therefore, we use two transformation matrices to jointly project these two features into the only one classification label space to get final classification result outputs. Given the two features $XZ$ and $LZ$, the optimization problem of our method can be rewritten as:
\begin{equation}\label{eq}
\begin{array}{c}
\min\limits_{W_1,W_2}\left\|Y-W_1XZ-W_2LX\right\|_F^2+\lambda_3\left\|W_1\right\|_F^2+\lambda_4\left\|W_2\right\|_F^2\\
\end{array}
\end{equation}
where $W_1\in \mathcal{R}^{c\times m}$ and  $W_2\in \mathcal{R}^{c\times m}$ are transformation matrices which process the corresponding main feature $XZ$ and the salient feature $LZ$ respectively. The remaining two $\lambda$s are positive regularization parameters. Through the above formula, we apply the extracted two features to the classification process, and classify the original data from two directions. On the one hand, the features extracted from the perspective of the column space contain the relationship information between the samples, and on the other hand, the features extracted from the row space are the most prominent features of the sample. And two corresponding transformation matrices are introduced to process the two features separately, so that the original single transformation matrix is relaxed. The regression tasks from different directions are respectively assigned to different transformation matrices, and the tasks of the original matrix are shared. Finally, the two transformation matrices jointly project the corresponding features into the classification space to obtain the final unified classification result, so that the features obtained from both the row and column spaces serve for the final classification.

\subsection{Solution to the Optimization Problem}
There are two unknown variables in the optimization problem eq.(\ref{eq}), which cannot be directly optimized. In order to obtain the optimal solution of the proposed algorithm, an iterative method is used solve the optimization problem in this subsection. When we update one variable, we fix another one. Here, the optimization method of eq.(\ref{eq}) with respect to $W_1$ and $W_2$ is presented as follows:

When fixing $W_2$, we can obtain:
\begin{equation}\label{eqop1}
W_1=(Y(XZ)^T-W_2 LX(XZ)^T)T_1
\end{equation}

When fixing $W_1$, we can obtain :
\begin{equation}\label{eqop2}
W_2=(Y(LX)^T-W_1 XZ(LX)^T)T_2
\end{equation}
where $T_1=(XZ(XZ)^T+\lambda_3 I)^{-1}$,  $T_2=(LX(LX)^T+\lambda_4 I)^{-1}$. The two features $XZ$ and $LZ$ are obtained from LatLRR. In order to speed up the calculation process, items $T_1$ and $T_2$ are unrelated to $W_1$ and $W_2$ which can be pre-calculated and used as a fixed item in the iterative process.

In summary, the process of our method is summarized in Algorithm 1.

\begin{flushleft}
	\resizebox{\textwidth}{!}{
	\begin{tabular}{p{1pt} p{12cm}}
		\hline
		\multicolumn{2}{l}{\textbf{Algorithm 1}: The proposed model} \\ \hline
		&\textbf{Input}: Training samples matrix $X$; binary label matrix $Y$; parameters $\lambda_3$, $\lambda_4$\\
		&\textbf{Initialization}: $W_1$= \textbf{0}, $W_2$= \textbf{0}, applying \textbf{LatlRR} method to get the principal feature $XZ$ and the salient feature $LX$ \\
		&1. Precalculating $T_1=(XZ(XZ)^T+\lambda_3 I)^{-1}$,  $T_2=(LX(LX)^T+\lambda_4 I)^{-1}$;\\
		&2. \textbf{While} not converged do \\
		&\quad 3. \textbf{Update} $W_1$ using eq.(\ref{eqop1}) \\
		&\quad 4. \textbf{Update} $W_2$ using eq.(\ref{eqop2}) \\
		&5. \textbf{End while} \\
		&\textbf{Output}: transformation matrices $W_1$ and $W_2$ \\ \hline
	\end{tabular}}
\end{flushleft}
\vspace{0.5cm}

\subsection{Classification}
In order to classify a testing sample $x_t\in \mathcal{R}^{m\times 1}$, there are three steps with our method: first, using LatLRR to extract features. When problem eq.(7) is solved, we obtain the matrices $L$ and $Z$. For the matrix $L$, we can directly obtain the salient feature $Lx_t$ of testing sample. For the matrix $Z$, we use the low rank projection method (\cite{dlrr}) to obtain the projection matrix $P$ that extracts the principal features of the testing sample $x_t$.
\begin{equation}
\min\limits_{P} \left\|P\right\|_*\quad s.t. PX=XZ
\end{equation}

By solving the above equation, a low rank projection matrix $P$ is obtained. The salient feature $Lx_t$ and the principal feature $Px_t$ of the test sample $x_t$ are thus obtained.

Second, we use these two features to directly apply $W_1$ and $W_2$ to get the projection result $W_1Px_t+W_2Lx_t$ of the test sample $x_t$.

Finally, we apply the nearest neighbor (NN) classifier to classify the projection result. For other classifiers, the results may be improved but is more involved. We leave it as a future work.

%\subsection{Computational Complexity Analysis}
%
%From the proposed algorithm (summarized in Algorithm 2), it mainly includes two parts, the first part calculates the main features and salient features to prepare for the subsequent transformation matrices learning, and the second part is to solve the two corresponding transformation matrices. Since the matrix addition, subtraction and multiplication are comparatively simple, the computational cost of these operations can be ignored. The main computational cost of LatLRR in the first part is the calculation of the SVD of matrices, thus the complexity of LatLRR is $O(m^2n+m^3)$ (assuming $X$ is $m\times n, m\leq n$) (\cite{latlrr}). The main computational cost of the second part is the inverse operation of $T_1$ and $T_2$ matrix. Fortunately, these two matrices can be pre-calculated because their values do not change in iteration. The computational complexity of matrix inversion is $O(m^3)$ to a $m\times m$ matrix. So the overall complexity of the proposed algorithm is about $O(m^2n+m^3+2m^3)$.

\section{Experimental Results and Analysis}

In this section, we compare the proposed method with several regression based methods and representation based methods on three widely used face databases and object database to validate the effectiveness. For regression based methods, DSLR (\cite{dlsr}), ReLSR (\cite{relsr}), GReLSR (\cite{grelsr}), RLR (\cite{rlr}), RLSL (\cite{rlsl}), and DRR (\cite{drr}) are utilized for comparison. For representation based methods, SRC (\cite{src}) is based on sparse representation, CRC (\cite{crc}) and proCRC (\cite{proCRC}) are based on collaborative representation, and LatLRR (\cite{latlrr}) as well as SALPL (\cite{alpl}) is based on low-rank representation. And for dictionary learning method, LCKSVD (\cite{lcksvd}) and RA-DPL (\cite{radpl}) are used in our experiments. For each group of experiments in different database, all the methods are repeated 20 times and samples are divided randomly into training set and test set. Then the mean accuracies are reported for comparison.

%For SRC, CRC, and proCRC, the classification results are obtained by selecting the class with the smallest reconstruction error and the normalized smallest reconstruction error. For LatLRR and SALPL, the extracted salient feature LX and the low-dimension salient feature QX are feed into NN classifier separately to get the corresponding classification accuracy. And the dimension $d$ used in SALPL is set to number of the classes in databases. For RLSL method, we set the dimension of the latent space to 200. For dictionary based method, LCKSVD uses the learned classifier while RA-DPL uses the reconstruction error to classify the testing samples. In addition, all the regression based methods classify the projected results by feeding them into the NN classifier. 

\begin{figure}[!t]
	\centering
	\subfigure[Some samples in the AR datbase]{
		\label{Fig.2(a)}
		\begin{minipage}[b]{0.69\textwidth}
			\includegraphics[scale=0.8]{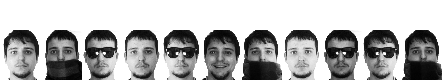} \\
			\includegraphics[scale=0.8]{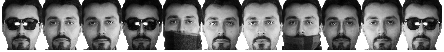}
		\end{minipage}
	}
	\subfigure[Some samples in the LFW datbase]{
		\label{Fig.2(b)}
		\begin{minipage}[b]{0.68\textwidth}
			\includegraphics[scale=0.99]{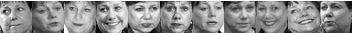} \\
			\includegraphics[scale=0.99]{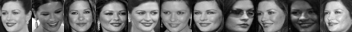}
		\end{minipage}
	}
	\subfigure[Some samples in the PIE datbase]{
		\label{Fig.2(c)}
		\begin{minipage}[b]{0.68\textwidth}
			\includegraphics[scale=0.375]{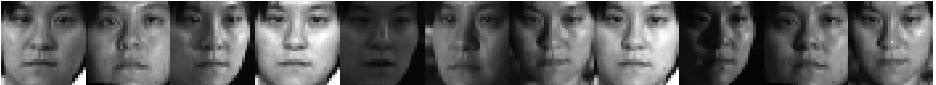} \\
			\includegraphics[scale=0.375]{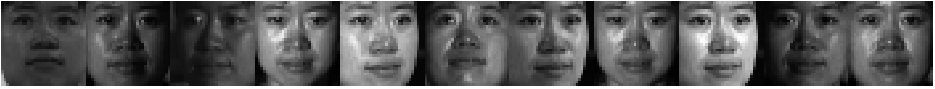}
		\end{minipage}
	}
	\caption{Some samples in different database}
	\label{Fig.2}
\end{figure}

\subsection{Experiments on Face Databases}
Four face databases, i.e., the AR database (\cite{ar}), the Labeled Faces in the Wild (LFW) database (\cite{lfw}), and the CMU Pose, Illumination, and Expression (CMU PIE) database (\cite{pie}) are adopted to evaluate different classification methods. Some samples in these databases are shown in Fig.\ref{Fig.2}. Detailed information about these three face databases is presented as follows:

\begin{enumerate}
	\item[(1)] The AR face database
\end{enumerate}

The original AR database contains over 4,000 color images corresponding to 126 people's faces (70 men and 56 women) with different facial expressions, illumination conditions, and occlusions (sun glasses and scarf) (\cite{ar}). We choose a subset which contains 3120 images of 120 people in the experiment. Each class has 26 images which were resized to $50\times 40$ in advance. We randomly select 5, 8, 10, 12, 15 images from each class as the training set while others as the test set. All the compared methods are repeated 20 times and the mean values of classification accuracies are reported in Table \ref{table1}.

Note that the bold numbers mean the best classification. We can see that our method is better than most of comparison methods when the number of selected training samples is small. When we further increase the training set, our method achieves the highest classification effect. These means that when the number of training sample of each class is small on the AR dataset, the classification information contained in the extracted low rank features is not enough. When the number increases, the extracted features are more robust, and our method also obtains the highest accuracies.

\begin{table}[htp]
	\small
	\centering
	\caption{Classification accuracies $(\%)$ of different methods on the AR database}
	\label{table1}
	\resizebox{\textwidth}{!}{
	\begin{tabular}{p{3.5cm}p{1.5cm}p{1.5cm}p{1.5cm}p{1.5cm}p{1.5cm}}
		\toprule
		Alg. & 5 & 8 & 10 & 12 & 15 \\
		\midrule
		SRC & 92.75 & 96.39 & 97.34 & 98.19 & 98.87 \\
		CRC & 92.21 & 95.85 & 96.85 & 97.87 & 98.47 \\
		proCRC & 91.38 & 95.25 & 96.19 & 96.06 & 97.87 \\
		SALPL & 91.61 & 96.21 & 97.36 & 98.11 & 98.84 \\
		LatLRR & 66.95 & 79.80 & 84.66 & 88.76 & 92.30 \\
		LCKSVD & 91.07 & 95.24 & 96.72 & 97.76 & 98.33 \\
		RA-DPL & 89.65 & 95.23 & 96.51 & 97.49 & 98.00 \\
		DLSR & 93.27 & 96.71 & 97.93 & 98.38 & 98.99 \\
		ReLSR & 93.11 & 96.95 & 97.71 & 98.19 & 98.71 \\
		GReLSR & 92.08 & 95.61 & 96.75 & 97.50 & 98.11 \\
		RLR & \textbf{94.60} & 96.73 & 97.62 & 98.25 & 98.61 \\
		RLSL & 92.34 & \textbf{97.45} & 97.70 & 98.39 & 98.86 \\
		DRR & 94.03 & 96.32 & 96.78 & 97.07 & 98.06 \\ \hline
		Our method & 94.29 & 97.25 & \textbf{98.12} & \textbf{98.54} & \textbf{99.03} \\ 
		\bottomrule
	\end{tabular}}
\end{table}

\begin{enumerate}
	\item[(2)] The LFW face database
\end{enumerate}

This database is designed for studying the problem of unconstrained face recognition. There are more than 13,000 faces images of 5749 people that are collected from the web in the original LFW face database (\cite{lfw}). A subset of original LFW database was adopted in our experiments, which contains 1251 images of 86 categories (\cite{lfw1251}). Each class has 11-20 images which are cropped and resized to $32\times 32$ pixels. We randomly select 2, 4, 6, 7, 8 images from each person and the rest for testing. All the compared methods are also repeated 20 times to obtain the mean values. The experimental results are shown in Table \ref{table2}.

\begin{table}[htp]
	\small
	\centering
	\caption{Classification accuracies $(\%)$ of different methods on the LFW database}
	\label{table2}
	\resizebox{\textwidth}{!}{
	\begin{tabular}{p{3.5cm}p{1.5cm}p{1.5cm}p{1.5cm}p{1.5cm}p{1.5cm}}
		\toprule
		Alg. & 2 & 4 & 6 & 7 & 8 \\ 
		\midrule
		SRC & 21.30 & 31.06 & 37.00 & 39.58 & 41.95 \\
		CRC & 22.34 & 30.42 & 36.56 & 37.40 & 39.55 \\
		proCRC & 22.97 & 30.49 & 35.41 & 37.32 & 40.05 \\
		SALPL & 17.87 & 26.85 & 33.33 & 35.53 & 37.58 \\
		LatLRR & 15.20 & 20.65 & 23.28 & 24.24 & 24.96 \\
		LCKSVD & 19.30 & 25.51 & 30.18 & 32.94 & 33.41 \\
		RA-DPL & 19.49 & 27.25 & 32.27 & 33.68 & 35.26 \\
		DLSR & 20.18 & 28.97 & 34.84 & 37.43 & 39.65 \\
		ReLSR & 18.58 & 28.84 & 35.15 & 37.23 & 39.28 \\
		GReLSR & 19.38 & 28.27 & 36.46 & 38.38 & 40.60 \\
		RLR & 17.38 & 28.56 & 36.37 & 38.35 & 39.39 \\
		RLSL & 22.34 & 32.97 & 38.91 & 41.60 & 43.69 \\
		DRR & 20.38 & 28.09 & 32.17 & 33.60 & 35.03 \\ \hline
		Our method & \textbf{23.67} & \textbf{33.80} & \textbf{40.37} & \textbf{42.16} & \textbf{44.86} \\ 
		\bottomrule
	\end{tabular}}
\end{table}

Obviously, on the LFW dataset, our method has achieved the best results regardless of the dataset size. Especially on the basis of the two features extracted by LatLRR, it is obvious that the classification accuracy obtained by using the two transformation matrices to process the feature information separately is greatly improved. And our method accuracy is about at least five percentages higher than the two methods that only use the features obtained from one angle, which just shows the effect of the sample relationship information contained in this data set on the classification (As mentioned above, the feature $XZ$ extracted by LatLRR exactly contains this type of information).

Compared to other regression-based methods, our method also achieves the highest classification accuracy. The difficulty in classifying samples of the LFW database is that they are acquired under different conditions, and contain many complicated scene pose and device difference information. It greatly increases the difficulty of classification, and as such, the sample contains a large amount of redundant information.  The reason why our proposed method performs well on this data set is that the feature extraction method processes this redundant information and also separates the complex information. At the same time, the two transformation matrices introduced also share the responsibility of a single matrix in this case. Moreover, LatLRR not only extracts the common features of the same class samples, but also contains information that is common to all samples and not related to the classification. Our method uses the label information and the transformation matrix to filter these features, and only selects features that are useful for classification, thus the transformed results is robust to some extent.

\begin{enumerate}
	\item[(2)] The CMU PIE face database
\end{enumerate}

CMU PIE database contains 41368 images of 68 people under 13 different poses, 43 different illumination conditions, and with 4 different expressions. We choose a subset of this database which contains 11554 images under five frontal pose, all the illustration and all the expression (\cite{pie}). There are 170 images for each person, and the size of each images is $32\times 32$ pixels after images are resized. For this database, we randomly select 5, 10, 20, 25, 30 images from each category and the rest of samples for testing.

\begin{table}[htp]
	\small
	\caption{Classification accuracies $(\%)$ of different methods on the CMU PIE database}
	\label{table3}
	\resizebox{\textwidth}{!}{
	\begin{tabular}{p{3.5cm}p{1.5cm}p{1.5cm}p{1.5cm}p{1.5cm}p{1.5cm}}
		\toprule
		Alg. & 5 & 10 & 20 & 25 & 35 \\
		\midrule
		SRC & 72.77 & 86.74 & 93.52 & 94.78 & 96.02 \\
		CRC & 73.05 & 86.07 & 92.89 & 94.20 & 94.69 \\
		proCRC & 76.72 & 88.65 & 94.12 & 94.99 & 95.72 \\
		SALPL & 73.14 & 86.84 & 91.04 & 92.43 & 93.53 \\
		LatLRR & 35.02 & 53.60 & 74.88 & 80.44 & 87.47 \\
		LCKSVD & 69.24 & 82.69 & 91.21 & 93.10 & 94.81 \\
		RA-DPL & 72.35 & 85.85 & 92.33 & 94.09 & 95.39 \\
		DLSR & 76.14 & 88.12 & 93.71 & 94.55 & 95.74 \\
		ReLSR & 76.12 & 88.07 & 93.45 & 94.57 & 95.67 \\
		GReLSR & 75.94 & 87.57 & 93.07 & 94.27 & 95.39 \\
		RLR & \textbf{81.10} & \textbf{89.95} & 93.84 & 94.57 & 95.50 \\
		RLSL & 75.80 & 87.44 & 93.15 & 95.04 & 95.59 \\
		DRR & 78.78 & 87.29 & 93.65 & 94.93 & 96.33 \\\hline
		Our method & 76.36 & 88.97 & \textbf{94.48} & \textbf{95.37} & \textbf{96.43} \\ 
		\bottomrule
	\end{tabular}}
\end{table}

All methods were repeated 20 times and the average accuracy was shown in Table \ref{table3}. Similar to the AR dataset, our method is less accurate than a few methods such as RLR and DRR when the training set is small, but as the training set increases, our accuracy rate becomes the highest. This also shows that our method requires a sufficient number of training samples to obtain enough generalized features on data sets like CMU PIE and AR.  If the training sample is insufficient, it may result in the extracted features being only applicable to a small number of training samples, but has no representative in the other block of samples.

Moreover, it's worth noting that on this dataset, our approach is always better than the method of using only the salient features in the classification stage. As a supervised improvement algorithm for LatLRR, SALPL performs much better than LatLRR on small training sample datasets. We introduce two transformation matrices to process row and column spatial information separately, and the classification accuracy is improved.

\subsection{Experiments on Object Databases}
In this subsection, all the classification methods are compared on the Columbia Object Image Library (COIL20) database (\cite{coil20}). Fig.\ref{Fig.3} lists some samples in the COIL20 database. The experimental results are given as follows:

\begin{figure}
	\centering
	\includegraphics[scale=1]{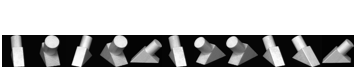}
	\includegraphics[scale=1]{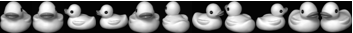}
	\caption{Some samples in COIL20 database}
	\label{Fig.3}
\end{figure}

The COIL20 dataset is a collection of images of 20 objects taken from different angles. There are 72 images per object. The data set contains two subsets. The first group contains images of five objects that contain both the object and the background. The second group contains images for all of the objects in which the background has been discarded (\cite{coil20}). In our experiments, the latter subset is used for comparing different methods. And each image in the subset is resized to $32\times 32$ pixels. Similarity, we randomly select 5, 10, 15, 20, 25 images from each category and the remainder for testing.

All methods were repeated 20 times on the COIL20 data set, and the average classification accuracies are shown in Table \ref{table4}.

It can be known from the experimental results that the proposed method performs better than other methods under different training set sizes. Another interesting finding is that the low-rank based method performs well on this dataset, even though the accuracy of LatLRR is better than some methods based on linear regression in some cases. The reason may be that each class of the COIL data set is a set of different objects at different angles, and the same type has a high correlation, so the low rank based method performs well on this data set.

\begin{table}[htp]
	\small
	\centering
	\caption{Classification accuracies $(\%)$ of different methods on the COIL20 database}
	\label{table4}
	\resizebox{\textwidth}{!}{
	\begin{tabular}{p{3.5cm}p{1.5cm}p{1.5cm}p{1.5cm}p{1.5cm}p{1.5cm}}
		\toprule
		Alg. & 5 & 10 & 15 & 20 & 25 \\
		\midrule
		SRC & 79.64 & 88.87 & 91.89 & 94.79 & 96.02 \\
		CRC & 82.50 & 89.53 & 92.79 & 94.33 & 95.31 \\
		proCRC & 82.88 & 91.38 & 94.54 & 96.21 & 97.16 \\
		SALPL & 84.66 & 92.65 & 95.84 & 97.43 & 98.18 \\
		LatLRR & 82.93 & 91.04 & 94.12 & 96.31 & 97.69 \\
		LCKSVD & 79.34 & 88.70 & 92.39 & 94.00 & 94.27 \\
		RA-DPL & 82.36 & 90.56 & 92.54 & 94.95 & 95.78 \\
		DLSR & 85.86 & 93.84 & 96.94 & 98.07 & 98.35 \\
		ReLSR & 84.44 & 91.81 & 95.10 & 96.22 & 97.43 \\
		GReLSR & 85.97 & 93.81 & 96.03 & 97.28 & 98.38 \\
		RLR & 86.04 & 92.35 & 95.15 & 96.81 & 97.47 \\
		RLSL & 85.46 & 93.88 & 96.20 & 97.10 & 97.76 \\
		DRR & 86.87 & 90.81 & 93.85 & 95.67 & 96.17 \\\hline
		Our method & \textbf{86.34} & \textbf{94.34} & \textbf{97.19} & \textbf{98.17} & \textbf{98.59} \\
		\bottomrule
	\end{tabular}
}
\end{table}

\begin{table*}
	\small
	\centering
	\caption{Classification accuracies $(\%)$ of different number of features and transformation matrices}
	\label{table5}
	\resizebox{\textwidth}{!}{
	\subtable{
		\label{} 
		\begin{tabular}{p{2.5cm}p{0.8cm}p{0.8cm}p{0.8cm}p{0.8cm}p{0.8cm}|p{0.8cm}p{0.8cm}p{0.8cm}p{0.8cm}p{0.8cm}}
			\toprule
			\multirow{2}{*}{different cases} & \multicolumn{5}{c|}{AR} & \multicolumn{5}{c}{LFW} \\ \cline{2-11}
			& 5 & 8 & 10 & 12 & 15 & 2 & 4 & 6 & 7 & 8 \\
			\midrule
			$y-w_2 lx$ & 93.77 & 96.87 & 97.97 & 98.52 & 99.01 & 20.48 & 30.25 & 35.24 & 37.04 & 38.82 \\
			$y-w(xz+lx)$ & 93.67 & 97.00 & 98.05 & 98.53 & \textbf{99.05} & 21.23 & 31.79 & 38.71 & 40.72 & 43.28 \\
			$y-w_1 xz-w_2 lx$ & \textbf{94.29} & \textbf{97.25} & \textbf{98.12} & \textbf{98.54} & 99.03 & \textbf{23.67} & \textbf{33.80} & \textbf{40.37} & \textbf{42.16} & \textbf{44.86} \\
			\bottomrule
		\end{tabular}
	}}
    
    \resizebox{\textwidth}{!}{
	\subtable{
		\label{} 
		\begin{tabular}{p{2.5cm}p{0.8cm}p{0.8cm}p{0.8cm}p{0.8cm}p{0.8cm}|p{0.8cm}p{0.8cm}p{0.8cm}p{0.8cm}p{0.8cm}}
			\toprule
			\multirow{2}{*}{different cases} & \multicolumn{5}{c|}{CMU PIE} & \multicolumn{5}{c}{COIL20} \\ \cline{2-11} 
			& 5 & 10 & 20 & 25 & 35 & 5 & 10 & 15 & 20 & 25 \\ 
			\midrule
			$y-w_2 lx$ & 76.64 & 88.55 & 94.09 & 94.97 & 96.09 & \textbf{86.91} & 93.33 & 95.93 & 97.13 & 97.90 \\
			$y-w(xz+lx)$ & 75.99 & 88.24 & 93.88 & 94.87 & 95.95 & 85.82 & 93.48 & 95.13 & 96.93 & 97.31 \\
			$y-w_1 xz-w_2 lx$ & \textbf{76.36} & \textbf{88.97} & \textbf{94.48} & \textbf{95.37} & \textbf{96.43} & 86.34 & \textbf{94.34} & \textbf{97.19} & \textbf{98.17} & \textbf{98.59} \\
			\bottomrule
		\end{tabular}
	}}
\end{table*}

\subsection{Experiments with Different Number of Features and Transformation Matrices}
In our method we use two parameters and two corresponding transformation matrices to accomplish the classification task together. To illustrate the role of the two parameters and matrices, in this section, we use different number of features and transformation matrices on different data sets and compare them to our proposed method. In the first case, only one salient feature $lx$ and the corresponding transformation matrix $w_2$ are used, and $y-w_2lx$ in Table \ref{table5} represents this case. And in the second case, two features, i.e., $xz$ and $lx$, are used but only one transformation matrix $w$ is used to process them, which is denoted by $y-w(xz+lx)$ in Table \ref{table5}. The last one is our method, which uses two features and corresponding transformation matrices, indicated by $y-w_1xz-w_2lx$. The final experimental results are shown in Table \ref{table5}.

From Table \ref{table5} we can see that whether using only one feature or using two features but only one transformation matrix, the accuracy rate is in most cases less than the two corresponding matrices we use. This difference is more prominent in the case of more complex sample conditions. It can be seen that using only one significant feature on the LFW dataset has a greater impact on the accuracy of the result than only one transformation matrix. This is because the sampling conditions of the LFW dataset are complex, and the classification lacks guidance of the inter-sample relationship information contained in the main features, resulting in classification errors. On the COIL20 dataset, it can be seen that when the number of training samples is 5, the accuracy rate is the highest when only one significant feature is used, but the accuracy of using the two features is worse. It is due to the fact that when the training samples are small, the principal features do not have good access to the inter-sample correlation information, thus adversely affect the final accuracy. In the AR dataset, the difference between several cases is very small. The experimental result is due to the subset of AR datasets used in the experiment, which is 50$\times$40. Due to the high dimensionality, the sample contains sufficient structural information. In addition, as can also be seen from Fig\ref{Fig.2(a)}, the samples in the AR data set do not have too much complicated information except for the scarf and sunglasses occlusion. In this case, even a single conversion matrix can give a good result. 

\subsection{Parameter Sensitivity and Selection}
There are four parameters that affect the experimental results , which are the parameters $\lambda_1$ and $\lambda_2$ used to balance the importance of the two features extracted by LatLRR, and the parameters $\lambda_3$ and $\lambda_4$ used to avoid the trivial solution of the two transformation matrices in our method. These four parameters are meaningful for learning the appropriate transformation matrices and getting projection results that contribute to the classification.

\begin{figure}[htp]
	\centering
%	\subfigure[Classification accuracy versus $\lambda_3$ and $\lambda_4$ on the AR datbase]{
%		\label{Fig.4(a)}
%		\begin{minipage}[b]{0.45\textwidth}
%			\includegraphics[scale=0.45]{para_ar_3D-eps-converted-to}
%		\end{minipage}
%	}
%	\subfigure[Classification accuracy versus $\lambda_3$ and $\lambda_4$ on the CMU PIE datbase]{
%		\label{Fig.4(b)}
%		\begin{minipage}[b]{0.5\textwidth}
%			\includegraphics[scale=0.45]{para_pie_3D-eps-converted-to}
%		\end{minipage}
%	}
	\subfigure[Classification accuracy versus $\lambda_3$ and $\lambda_4$ on the LFW datbase]{
		\label{Fig.4(c)}
		\begin{minipage}[b]{0.45\textwidth}
			\includegraphics[scale=0.35]{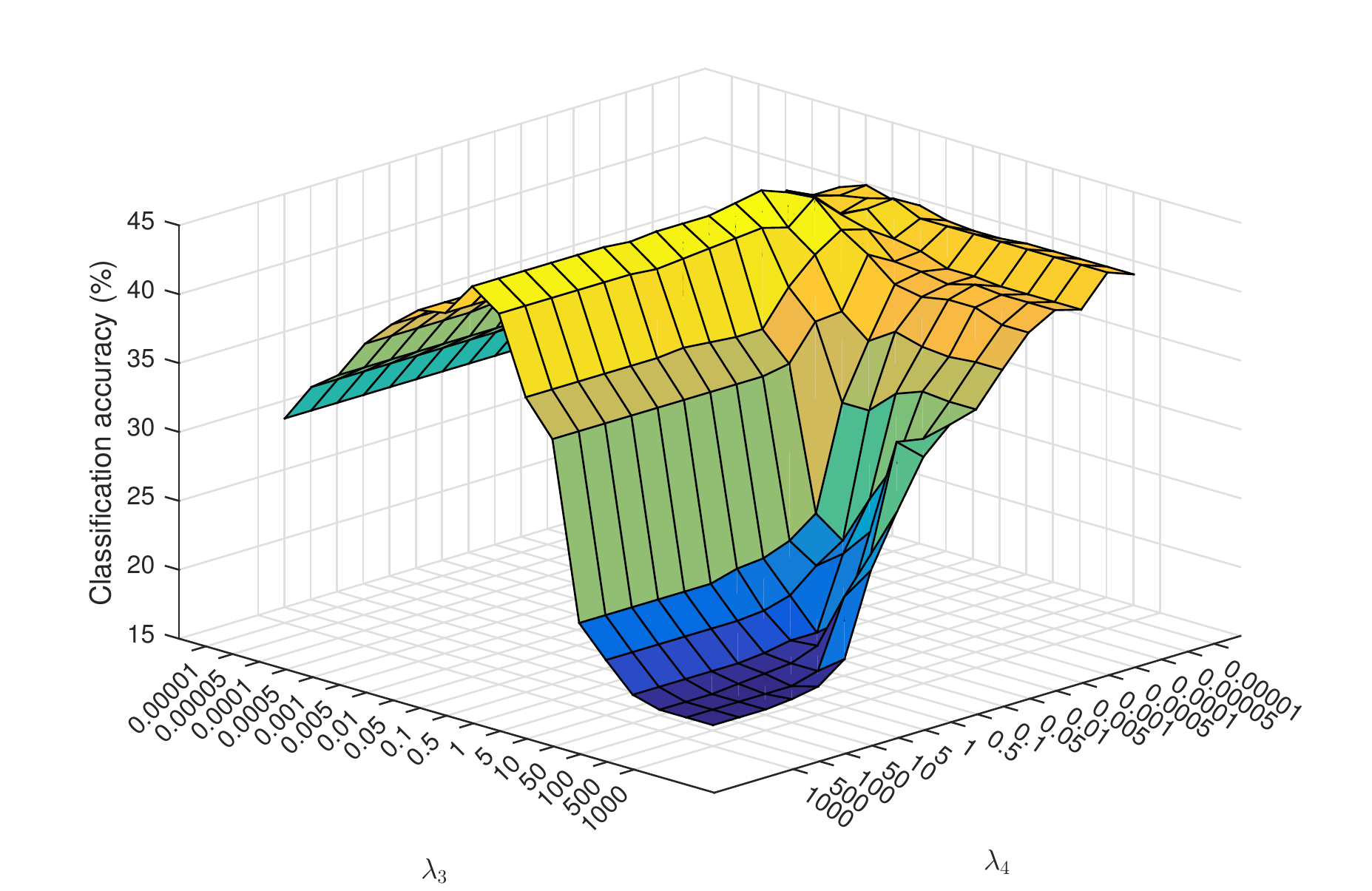}
		\end{minipage}
	}
	\subfigure[Classification accuracy versus $\lambda_3$ and $\lambda_4$ on the COIL datbase]{
		\label{Fig.4(d)}
		\begin{minipage}[b]{0.5\textwidth}
			\includegraphics[scale=0.33]{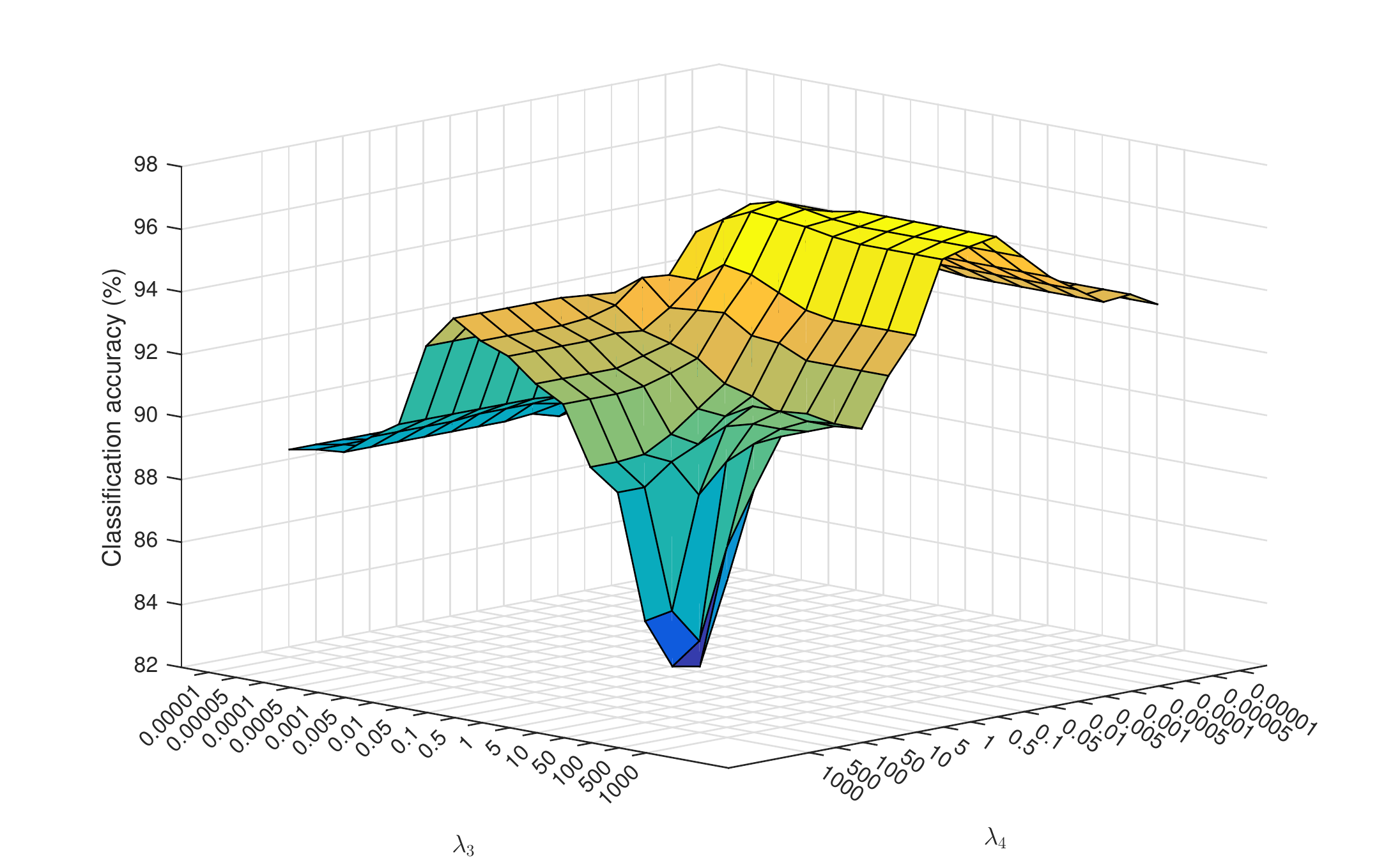}
		\end{minipage}
	}
	\caption{Relationships of the classification accuracy (\%) and different combinations of parameters on the different datasets.
		%		, (a) The AR dataset, in which 10 samples of each class are selected as the training samples (where $\lambda_1=10$, $\lambda_2=0.5$). (b) The CMU PIE dataset, in which 20 samples of each class are selected as the training samples (where $\lambda_1=100$, $\lambda_2=0.5$)
		(a) The LFW dataset, in which 8 samples of each class are selected as the training samples (where $\lambda_1=1$, $\lambda_2=5$). (b) The COIL dataset, in which 10 samples of each class are selected as the training samples (where $\lambda_1=0.05$, $\lambda_2=1$).}
	\label{Fig.4}
\end{figure}

In order to adjust these four parameters, we set a parameter candidate set to $\{1e^{-5},5e^{-5},1e^{-4},5e^{-4},...,5e^2,1e^3\}$, and apply different combinations of parameters to the algorithm. The adjustments of $\lambda_1,\lambda_2$ and $\lambda_3,\lambda_4$ are done separately. In the preprocessing stage of LatLRR, various combinations of $\lambda_1$ and $\lambda_2$ are applied to the feature extraction algorithm (\cite{latlrr}). After obtaining $\lambda_1$ and $\lambda_2$, their values are fixed. Then $\lambda_3$ and $\lambda_4$ are adjusted, and the the relationships of the classification accuracy and the two parameters on the LFW and COIL20 databases in Fig.\ref{Fig.4}.

From Fig.\ref{Fig.4}, we can find that the classification accuracy changes differently on the different datasets, which means that the extracted principal features and salient features contribute differently to the classification on different data sets. It can be observed that the proposed method obtains satisfactory effect when $\lambda_3$ and $\lambda_4$ respectively locate in the range of $[5e^{-2},1e^{0}]$ and $[1e^{-3},1e^{-1}]$ on the datasets given in Fig.\ref{Fig.4}. Through tuning $\lambda_3$ and $\lambda_4$, the optimal combination of them can be found.

From above separate adjustments of $\lambda_1,\lambda_2$ and $\lambda_3,\lambda_4$, thus the optimal combination of the four parameters can be obtained, and it is used as a parameter to calculate 20 times on the given database, and the average classification accuracy is taken for comparison.

\begin{figure}[htp]
	\subfigure[AR]{
		\label{Fig.5(c)}
		\begin{minipage}[b]{0.467\textwidth}
			\centering
			\includegraphics[scale=0.45]{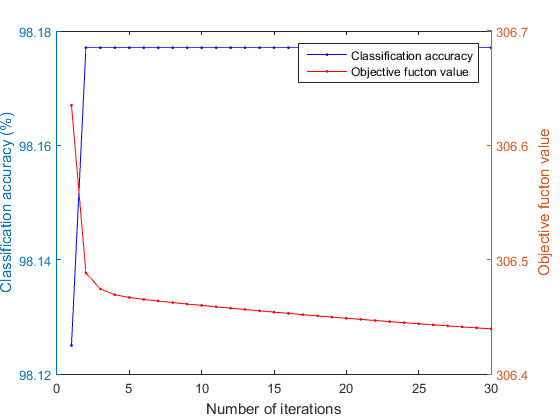}
		\end{minipage}
	}
	\subfigure[CMU PIE]{
		\label{Fig.5(b)}
		\begin{minipage}[b]{0.5\textwidth}
			\centering
			\includegraphics[scale=0.45]{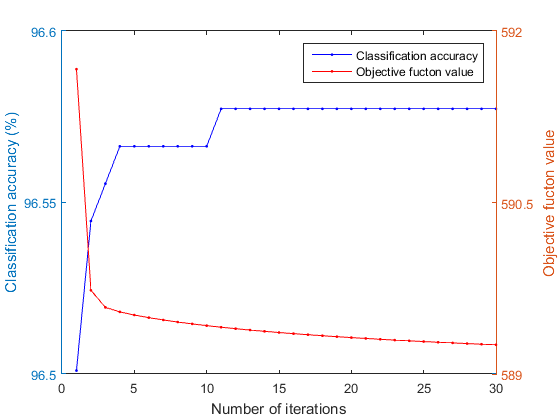}
		\end{minipage}
	}
	\subfigure[LFW]{
		\label{Fig.5(a)}
		\begin{minipage}[b]{0.467\textwidth}
			\centering
			\includegraphics[scale=0.45]{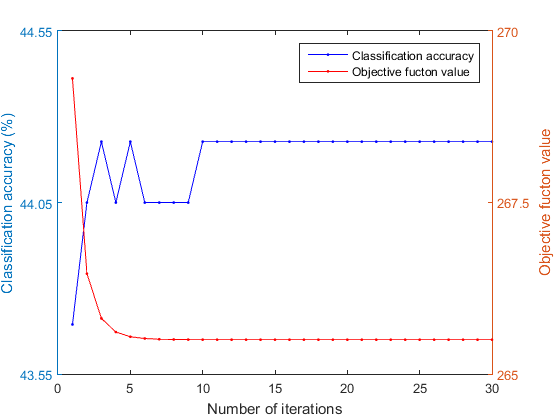}
		\end{minipage}
	}
	\subfigure[COIL]{
		\label{Fig.5(d)}
		\begin{minipage}[b]{0.5\textwidth}
			\centering
			\includegraphics[scale=0.45]{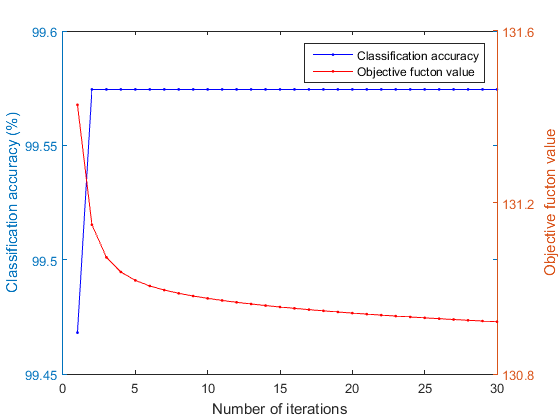}
		\end{minipage}
	}
    \caption{Objective function value and classification accuracy (\%) versus the number of iterations of the proposed method on the (a) AR (b) CMU PIE (c) LFW (d) COIL datasets, in which 10, 35, 8 and 25 samples of each class are randomly selected as the training samples, respectively.}
	\label{Fig.5}
\end{figure}

\subsection{Convergence Study}
We plot the curves of objective function values and classification accuracies on the used datasets in our paper with respect to the number of iterations in Fig.\ref{Fig.5}. The objective function value is calculated by $\tfrac{1}{2}\left\|Y-W_1XZ-W_2LX\right\|_F^2\\+\tfrac{\lambda_3}{2}\left\|W_1\right\|_F^2+\tfrac{\lambda_4}{2}\left\|W_2\right\|_F^2$. It is obvious that the objective value monotonically decreases as the number of iterations increases, and gradually reaches a stable value. We can also see that the optimization algorithm converges fast within a few iterations. Our method can find the local optimal value quickly. The curve of the classification accuracies in Fig.\ref{Fig.5} shows that the accuracy rate obtains stable values as the number of iterations increases, which also illustrates the convergence of our method from another perspective.

\section{Conclusions and Discussions}
This paper proposes a new method of a double transformation matrices learning based on low rank features. Different from other methods that use a single transformation matrix to process the original sample, the proposed method uses double transformation matrices to process the low-rank features extracted from the original sample data. The above observation results show that the optimization algorithm of solve the information redundancy problem of samples. The two features obtained by feature extraction respectively contain the salient features of the sample and the relationship between samples, and then our method uses two transformation matrices to jointly project them onto the same label space, so that both features serve for the final classification. Our method exhibits the role of multiple features in the same classification task, and reduces the pressure on the transformation matrix in regression tasks. In our experiments, especially under more complex sample conditions, our experimental results demonstrate the effectiveness of the proposed method. However, we note that our method does not perform well in the case of insufficient training samples. This is because when the training samples are small, the features obtained by low-rank feature extraction may not be applicable to all samples. Moreover, due to the low-rank features need to be extracted first, the amount of calculation at this step makes the classification slower than some methods. So the subsequent work plan is to try other feature extraction schemes. And the feature extraction and regression processes are separated from each other. The method may fall into a local optimum instead of a global optimum, therefore it is necessary to find global optimal solution of the problem in the future.

\section*{Acknowledgments}
This work is supported by the National Natural Science Foundation of China under Grant Nos. 61672265, U1836218, the 111 Project of Ministry of Education of China under Grant No. B12018.

\bibliography{mybibfile}
%\end{spacing}
\end{document}